\documentclass{article} 
\PassOptionsToPackage{table}{xcolor} 
\usepackage{iclr2027_conference,times}

\usepackage{amsmath,amsfonts,bm}

\def\eqref#1{equation~\ref{#1}}

\def\1{\bm{1}}

\DeclareMathAlphabet{\mathsfit}{\encodingdefault}{\sfdefault}{m}{sl}
\SetMathAlphabet{\mathsfit}{bold}{\encodingdefault}{\sfdefault}{bx}{n}

\usepackage[table]{xcolor}
\definecolor{AweAIblue}{HTML}{386B58}
\definecolor{AweAIgreen}{HTML}{529A69}
\usepackage[colorlinks=true, linkcolor=AweAIblue, citecolor=AweAIblue, urlcolor=AweAIblue, hypertexnames=false]{hyperref}
\usepackage{url}
\usepackage[utf8]{inputenc} 
\usepackage[T1]{fontenc}    
\usepackage{booktabs}       
\usepackage{listings}
\usepackage[most]{tcolorbox}
\usepackage{threeparttable}
\usepackage{tabularx}
\usepackage{ragged2e}
\usepackage{amsfonts}       
\usepackage{nicefrac}       
\usepackage{arydshln}
\usepackage{microtype}      
\usepackage{graphicx}
\usepackage{mathrsfs}
\usepackage{amsmath}

\usepackage{footnote}
\usepackage{footnotebackref}
\usepackage{multirow}
\usepackage{makecell}
\usepackage{algorithm}
\usepackage{algorithmic}

\usepackage{float}  
\usepackage{tikz}
\usetikzlibrary{shapes,arrows}
\usepackage{appendix}
\usepackage{bm}
\usepackage{amsbsy}
\usepackage{pifont}
\usepackage{wrapfig}
\usepackage{bbding}
\usepackage{titlesec}
\usepackage{colortbl}
\usepackage{subcaption}
\usepackage{etoc}     
\usepackage{enumitem}

\newcommand{\dinglabelone}{\textcolor[HTML]{386B58}{\ding{182}}}
\newcommand{\dinglabeltwo}{\textcolor[HTML]{386B58}{\ding{183}}}
\newcommand{\dinglabelthree}{\textcolor[HTML]{386B58}{\ding{184}}}

\definecolor{S2Sage}{HTML}{4F8071}
\definecolor{S2Paper}{HTML}{F7F3EA}
\definecolor{S2Line}{HTML}{D6DED7}
\definecolor{AblationBlue}{HTML}{4F8FCC}
\definecolor{AblationGreen}{HTML}{4FA66E}
\definecolor{AblationPink}{HTML}{D85B62}

\newtcblisting{PromptBox}[2][]{
    enhanced jigsaw,
    breakable,
    listing only,
    listing options={
        basicstyle=\footnotesize\ttfamily,
        breaklines=true,
        breakatwhitespace=false,
        columns=fullflexible,
        keepspaces=true,
        aboveskip=0pt, belowskip=0pt,
        breakindent=0pt,
        breakautoindent=false,
        upquote=true,
        literate=
            {"}{\char`\"}1
            {'}{\char`\'}1
            {"}{\char`\"}1
            {'}{\char`\'}1
            {…}{...}1
            {—}{---}1
            {–}{--}1,
    },
    colframe=S2Line,
    colback=S2Paper,
    coltitle=white,
    colbacktitle=S2Sage,
    title={#2},
    fonttitle=\large\bfseries,
    arc=4mm,
    boxsep=3pt,
    top=2pt, bottom=2pt,
    #1
}

\title{SCBO: Semantically Coherent Batching and Ordering for LLM-Based Social Surveys}

\author{Yuanzi Li, Lingjie Wang, Zihang Tian, Lei Wang, Xu Chen\thanks{Corresponding author.}\\
Renmin University of China\\
{\fontfamily{pcr}\selectfont liyuanzi0313@outlook.com, xu.chen@ruc.edu.cn}
}

\newcommand{\model}{SCBO}

\iclrfinalcopy 
\begin{document}

\maketitle

\ificlrfinal
    \lhead{Published as a conference paper at ICLR 2027}
\else
    \lhead{Under review as a conference paper at ICLR 2027}
\fi

\begin{abstract}
Large Language Models (LLMs) offer a scalable way to simulate survey respondents conditioned on demographic profiles and observed reference responses. However, the conventional one-question-per-prompt paradigm is limited in three respects: it repeatedly encodes the same context, incurring substantial token and inference costs; it restricts each target to its own narrow subset of observed responses, preventing reference evidence from being shared across targets; and it predicts every answer in isolation, preventing later predictions from leveraging information in earlier answers. To address these limitations, we predict multiple questions in a single prompt, which amortizes the shared context, allows multiple targets to share a broader pool of observed reference responses, and enables later predictions to condition on earlier ones. However, such a method faces two challenges: (1) how to form semantically coherent batches and select shared references, and (2) how to order questions and references to improve autoregressive generation. We propose Semantically Coherent Batching and Ordering (SCBO), a training-free framework that addresses these challenges through two modules, preceded by a Question Instantiation step in which an LLM extracts compact semantic representations from survey items to filter out template noise. (1) Semantic Batching and Selection: SCBO groups related questions into semantic batches and constructs a shared reference bank by combining target-specific retrieval with centroid-based completion. (2) Curriculum-based Ordering: It then applies a heuristic easy-to-hard ordering to target questions and orders references within the shared bank according to their semantic alignment with the ordered questions. Experiments on four large-scale survey datasets and four LLMs show that SCBO substantially reduces token consumption and inference time while generally improving prediction accuracy over the non-batched baseline.
Code is available at \url{https://anonymous.4open.science/r/SCBO-41D8}.
\end{abstract}
\section{Introduction}

Social surveys, which ask people directly about their attitudes, values, and political preferences, are a foundational tool in the social sciences for tracking public opinion and evaluating policy \citep{groves2011survey, roopa2012questionnaire, tourangeau2000psychology, squazzoni2020computational, hamill2009social, wang2025user}. However, asking real people is hard: respondents are costly to recruit, may not answer honestly, and often refuse the most revealing questions, such as income or political extremes \citep{wright2010survey, heffetz2019difficulty, kalton2009methods}. As a result, traditional surveys remain limited in scale, reliability, and coverage.

Large Language Models (LLMs) \citep{achiam2023gpt,schaeffer2023emergent, kosinski2023theory} offer an opportunity to alleviate these difficulties. Trained on vast amounts of human-written Internet text, which contains traces of how people express attitudes, judgments, and preferences, LLMs acquire rich prior knowledge of human attitudes and behavior \citep{argyle2023out, park2023generative, aher2023using, cao2023assessing, zhou2025should, durmus2023towards}. 
Recent work therefore uses LLMs as synthetic respondents: given a respondent's demographic profile and reference responses, a model can predict their responses to other survey questions, extending survey scale and coverage at a fraction of the effort \citep{santurkar2023whose,  simmons2023moral, bisbee2024synthetic}.

While LLM-based simulation relieves the burden of collecting answers from people, the prevailing ``one-question-per-prompt'' practice \citep{argyle2023out, santurkar2023whose, bisbee2024synthetic} is limited in three respects.
\emph{(1) It is expensive:} every prompt repeats the task instruction, the respondent's demographic profile, and a few of that respondent's observed answers as reference examples, so this shared context is re-encoded once per target question.
As shown in Table~\ref{tab:cost_estimation}, exhaustive prediction involves 25.0 million user-question pairs for WVS and 3.5 million for ANES, requiring 14.06B and 2.18B input tokens; for WVS alone, this lower-bound estimate costs at least \$11,126 on GPT-6-Astra, \$58,552 on Claude-Fable-5, and \$11,017 on Gemini-3.8-Flash.
\emph{(2) It prevents reference sharing:}
each target question independently retrieves $n$ references.
When targets are predicted separately, references retrieved for one target
are unavailable to the others, so each prediction uses only a small portion
of the respondent's observed responses.
\emph{(3) It discards information:} answering each question in a prompt of its own treats a respondent's answers as mutually independent, whereas items probing the same underlying attitude are strongly correlated, and the answer to one question shapes the answer to the next.
Each prediction is thus made without access to the answers that would inform it most.

We propose to batch target questions into a prompt: the shared context is encoded once per batch rather than once per question, questions within the same batch can share a larger pool of observed reference responses under the same context budget, and later predictions can condition on answers generated earlier in the batch.
However, answering questions together makes each prediction depend on which questions accompany it and in what order, giving rise to two challenges:
\textbf{\emph{Challenge 1: What to put in a batch.}} All questions in a batch share one set of references, and that set has a limited budget. If unrelated questions are grouped together, say one about trust in government and one about family values, examples that suit one are of little use to the other, and most questions end up with no useful reference. Batches therefore have to be formed so that a single set of examples can serve every question in them, and the examples have to be chosen to cover the whole batch rather than any one question.
\textbf{\emph{Challenge 2: How to order a batch.}} An LLM generates batched answers autoregressively, so later predictions can condition on answers generated earlier, but not vice versa. If an earlier question is predicted correctly, its answer can provide useful evidence for subsequent predictions. For example, a correct prediction of religious importance may help predict a respondent's view on same-sex marriage. Questions should therefore be ordered so that reliable earlier predictions support harder ones that follow, with references ordered to align semantically with the target sequence.

\definecolor{hdrgray}{RGB}{233,236,242}
\definecolor{scalecol}{RGB}{237,245,240}
\definecolor{tokencol}{RGB}{240,243,250}
\definecolor{costcol}{RGB}{252,239,235}
\definecolor{costtext}{RGB}{176,48,32}

\begin{table}[!t]
\caption{Estimated token usage and API cost for exhaustive
one-question-per-prompt prediction. Prices are based on the public rates listed at
\url{https://aiapi.world/pricing}.}
\vspace{-0.3em}
\label{tab:cost_estimation}
\centering
\footnotesize
\setlength{\tabcolsep}{4pt}
\renewcommand{\arraystretch}{1.16}
\setlength{\aboverulesep}{0pt}\setlength{\belowrulesep}{0pt}
\setlength{\extrarowheight}{0.28ex}
\resizebox{\textwidth}{!}{%
\begin{tabular}{l
  >{\columncolor{scalecol}}r >{\columncolor{scalecol}}r >{\columncolor{scalecol}}r
  >{\columncolor{tokencol}}r >{\columncolor{tokencol}}r >{\columncolor{tokencol}}r >{\columncolor{tokencol}}r
  >{\columncolor{costcol}}r >{\columncolor{costcol}}r >{\columncolor{costcol}}r}
\toprule
\rowcolor{hdrgray}
& & & &
\multicolumn{2}{c}{\cellcolor{hdrgray}\textbf{Avg. tokens}} &
\multicolumn{2}{c}{\cellcolor{hdrgray}\textbf{Token budget (M)}} &
\multicolumn{3}{c}{\cellcolor{hdrgray}\textbf{Estimated cost (USD)}} \\
\arrayrulecolor{black!45}\cmidrule(lr){5-6}\cmidrule(lr){7-8}\cmidrule(lr){9-11}\arrayrulecolor{black}
\rowcolor{hdrgray}
\multirow{-2}{*}{\textbf{Dataset}} & \multirow{-2}{*}{\textbf{Users}} & \multirow{-2}{*}{\textbf{Questions}} & \multirow{-2}{*}{\textbf{(User, Question)}} & \textbf{Context} & \textbf{Question} & \textbf{Input} & \textbf{Output} & \textbf{GPT-6} & \textbf{Fable-5} & \textbf{Gemini-3.8} \\
\midrule
WVS & 97,220 & 285 & 25,038,918 & 491.1 & 70.6 & 14,064 & 125 & \textcolor{costtext}{\textbf{\$11,126+}} & \textcolor{costtext}{\textbf{\$58,552+}} & \textcolor{costtext}{\textbf{\$11,017+}} \\
ANES & 5,521 & 952 & 3,541,672 & 551.4 & 62.9 & 2,176 & 18 & \textcolor{costtext}{\$1,714+} & \textcolor{costtext}{\$9,026+} & \textcolor{costtext}{\$1,698+} \\
GSS & 3,986 & 539 & 1,001,083 & 539.5 & 55.1 & 595 & 5 & \textcolor{costtext}{\$470+} & \textcolor{costtext}{\$2,473+} & \textcolor{costtext}{\$465+} \\
BSA & 4,120 & 251 & 531,064 & 509.5 & 63.2 & 304 & 3 & \textcolor{costtext}{\$240+} & \textcolor{costtext}{\$1,265+} & \textcolor{costtext}{\$238+} \\
\bottomrule
\end{tabular}%
}
\vspace{-1em}
\end{table}

To address these challenges, we propose a \textbf{Semantically Coherent Batching and Ordering (SCBO) Framework} that maximizes accuracy while reducing token consumption and inference time via two modules. \emph{(1) Semantic Batching and Selection:} To address Challenge 1, we utilize fixed-size clustering to group cohesive questions. Within each batch, we retrieve target-specific references into a shared reference bank and fill remaining slots with centroid-based retrieval when budget allows. \emph{(2) Curriculum-based Ordering:} To address Challenge 2, we exploit the autoregressive dependency among batched predictions and draw inspiration from curriculum learning. We place easier questions earlier, where their predictions are more likely to provide reliable context for harder questions that follow, and arrange references to align with the target sequence.

Our contributions are threefold:
\dinglabelone\ We formulate the LLM-based survey batching problem, which amortizes respondent context across questions, enables multiple targets to share a broader pool of observed reference responses, and allows later predictions to condition on earlier answers. We identify its two challenges: what to put in a batch and how to order what is in it.
\dinglabeltwo\   We propose \textbf{SCBO}, a training-free framework with two modules: semantic batching, which groups cohesive questions and selects a shared reference bank, and curriculum-based ordering, which arranges questions and references so that earlier predictions inform later ones.
\dinglabelthree\   Experiments on four large-scale survey datasets across four LLMs show that SCBO reduces token cost by over 50\% and accelerates inference by 3.3--14.0$\times$, while generally improving accuracy over the one-question-per-prompt baseline.

\section{Problem Formulation: LLM-based Social Surveys}
\label{sec:problem_formulation}

Let $u$ denote a respondent associated with a textual demographic profile $P_u$ (e.g., age, gender, education, and income). For this respondent we have a reference set $(\mathcal{Q}_{pool}, \mathcal{A}_{pool})$ of survey questions that $u$ has answered before, together with the given answers, which serve as candidate in-context examples. The task is to predict how $u$ would answer a target set $\mathcal{Q}_{target} = \{q_1, \dots, q_N\}$ of $N$ questions that $u$ has not been asked.

Under the prevailing one-question-per-prompt paradigm, an LLM $\mathcal{M}$ predicts each target answer from its own prompt. Let $\operatorname{card}(\cdot)$ denote the number of examples in a reference set:
\begin{equation}
\label{eq:single}
\hat{a}_i = \mathcal{M}\big(\mathcal{I} \oplus P_u \oplus S_i \oplus q_i \big),
\qquad
S_i \subset (\mathcal{Q}_{pool}, \mathcal{A}_{pool}),
\quad
\operatorname{card}(S_i) = n.
\end{equation}
where $\mathcal{I}$ is the task instruction, $S_i$ is a reference set retrieved for $q_i$ under a per-target-question budget $n$, and $\oplus$ denotes concatenation into a single prompt. Each prediction $\hat{a}_i$ is evaluated against the withheld ground-truth answer $a_i$ in terms of ACC, MAE, and F1 over the $N$ target questions.

Only $S_i$ and $q_i$ in Eq.~(\ref{eq:single}) vary across target questions, whereas $\mathcal{I}$ and $P_u$ are identical for all of them. Let $T_{\text{single}}$ denote the total number of prompt tokens needed to predict all $N$ target answers of one respondent under Eq.~(\ref{eq:single}). Writing $|\cdot|$ for the token count of a prompt component,
\begin{equation}
\label{eq:cost_single}
T_{\text{single}} = \sum_{i=1}^{N}\big(|\mathcal{I}| + |P_u| + |S_i| + |q_i|\big) = \underbrace{N\big(|\mathcal{I}| + |P_u|\big)}_{\text{redundant}} + \sum_{i=1}^{N}\big(|S_i| + |q_i|\big).
\end{equation}
This formulation is limited in three respects.
\emph{Cost:} the first term of Eq.~(\ref{eq:cost_single}) grows linearly in $N$ while carrying no new information, and it dominates in practice, as on WVS the shared context averages $491.1$ tokens per prediction against $70.6$ tokens for the target question itself (Table~\ref{tab:cost_estimation}).
\emph{History:} under a fixed reference budget,, each target question independently retrieves and uses its reference set $S_i$. Because they are used separately, each target can access only the portion of respondent history contained in its own set, even when references retrieved for other targets could also be informative.
\emph{Independence:} Eq.~(\ref{eq:single}) conditions $\hat{a}_i$ only on $q_i$ and the reference set, so the $N$ answers are predicted independently, without conditioning on answers generated for other target questions. Survey responses are not: items that probe the same underlying attitude are strongly correlated, and an answer given to one question shapes the answer given to the next.

To address all three limitations, we let one prompt answer several questions at once, partitioning $\mathcal{Q}_{target}$ into $K=N/C$ disjoint batches $\mathcal{B}=\{B_1,\dots,B_K\}$ of size $C$ and rewriting Eq.~(\ref{eq:single}) as
\begin{equation}
\label{eq:batched}
\{\hat{a}_i\}_{q_i \in B_k}
=
\mathcal{M}\big(
\mathcal{I}
\oplus P_u
\oplus \boldsymbol{\phi}(S_k)
\oplus \boldsymbol{\pi}(B_k)
\big),
\qquad
S_k \subset (\mathcal{Q}_{pool}, \mathcal{A}_{pool}),
\quad
\operatorname{card}(S_k)=m.
\end{equation}
where $S_k$ is a reference bank shared by all questions in $B_k$, with a budget of $m=nC$ examples, matching the aggregate reference budget of the $C$ individual target questions.
The functions $\boldsymbol{\pi}$ and $\boldsymbol{\phi}$ order the questions and references within the prompt. 
The corresponding token cost is
\begin{equation}
\label{eq:cost_batched}
T_{\text{batch}} = K\big(|\mathcal{I}| + |P_u|\big) + \sum_{k=1}^{K}\Big(|S_k| + \textstyle\sum_{q_i \in B_k}|q_i|\Big).
\end{equation}
Comparing Eq.~(\ref{eq:batched}) with Eq.~(\ref{eq:single}), the shared context $\mathcal{I} \oplus P_u$ now serves $C$ questions instead of one, the questions in each batch share a reference bank $S_k$ rather than using separate per-target sets $S_i$, and the answers of a batch are generated within an autoregressive sequence rather than in separate prompts. This addresses all three limitations. On cost, the dominant term drops from $N$ copies in Eq.~(\ref{eq:cost_single}) to $K = N/C$ in Eq.~(\ref{eq:cost_batched}), a reduction by a factor of $C$, while the questions are still encoded exactly once. On history, the shared bank allows each target to access references retrieved for other questions in the same batch, exposing it to a broader range of respondent history under the reference budget $m=nC$. On independence, later answers can condition on answers generated earlier in the batch, so the correlations among a respondent's answers can be exploited rather than ignored.

The core of Eq.~(\ref{eq:batched}) therefore lies in three quantities it leaves unspecified: the partition $\mathcal{B}$, the shared bank $S_k$, and the orderings $(\boldsymbol{\pi}, \boldsymbol{\phi})$. 
These respectively determine which target questions are grouped together, which historical responses are shared within each batch, and how the target questions and references are ordered.
All three decide what each answer is conditioned on, and hence the prediction accuracy. The next section specifies how SCBO determines them.

\section{The SCBO Framework}
\label{sec:methodology}

\begin{figure}[!t]
    \centering
    \includegraphics[width=\textwidth]{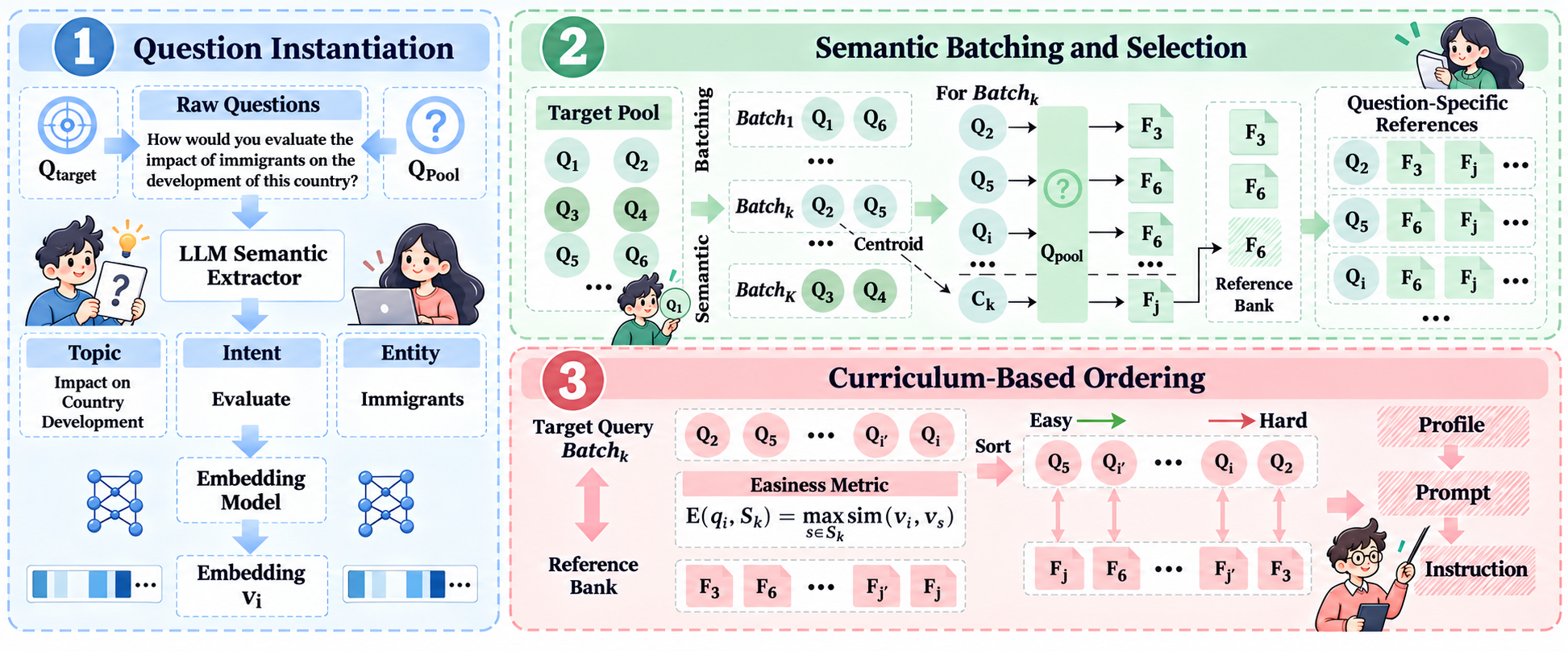}
    \vspace{-1em}
    \caption{Overview of SCBO. As preprocessing, Question Instantiation extracts compact topic, intent, and entity from survey questions to filter template noise for the two modules: (1) Semantic Batching clusters semantically cohesive questions and builds shared reference banks through target-specific and centroid-based retrieval; and (2) Curriculum-based Ordering sorts target questions from easy to hard and orders references within each shared bank by semantic alignment.}
    \vspace{-1em}

    \label{fig:scbo-framework}
\end{figure}

As illustrated in Figure~\ref{fig:scbo-framework}, \textbf{SCBO} determines the three quantities that Eq.~(\ref{eq:batched}) leaves unspecified. We first map every survey item to a compact semantic representation that strips template noise, a preprocessing step we call Question Instantiation (Section~\ref{subsec:instantiation}). Then, operating on these representations, Semantic Batching and Selection fixes the partition $\mathcal{B}$ and the shared reference bank $S_k$, which addresses Challenge 1 (Section~\ref{subsec:batching}). Finally, Curriculum-based Ordering fixes the orderings $\boldsymbol{\pi}$ and $\boldsymbol{\phi}$, which addresses Challenge 2 (Section~\ref{subsec:ordering}).

\subsection{Preprocessing: Question Instantiation for Semantic Denoising}
\label{subsec:instantiation}

Raw survey items often contain redundant conversational templates and standardized instructions that obscure the core semantic intent. To avoid embedding this noise, we extract the semantic essence of each question $q_i \in \mathcal{Q}$ using an LLM $\mathcal{M}$ with a prompt $\mathcal{P}_{extract}$, which distills a representation $\tilde{q}_i = \mathcal{M}(q_i, \mathcal{P}_{extract}) = \{T_i, I_i, E_i\}$, where $T_i$, $I_i$, and $E_i$ denote the Topic, Core Intent, and Key Entities, respectively; the prompt is given in Appendix~\ref{subsec:appendix_instantiation_prompt}. For instance, the raw item \emph{``How would you evaluate the impact of immigrants on the development of this country?''} is reduced to $T_i=$ \emph{impact on country development}, $I_i=$ \emph{evaluate}, and $E_i=$ \emph{immigrants}: the respondent-addressing frame \emph{``How would you''}, which recurs across items and contributes little to topical discrimination, is discarded, while the country-level scope expressed by \emph{``of this country''} is retained in the topic representation \emph{impact on country development}. An embedding model encodes each representation as $\mathbf{h}_i=f_{emb}(\tilde{q}_i)\in\mathbb{R}^d$. We then apply L2 normalization, $\mathbf{v}_i=\mathbf{h}_i/\|\mathbf{h}_i\|_2$. Subsequent clustering, retrieval, and ordering operations use these normalized embeddings. Consequently, cosine similarity is equivalent to the dot product, and the squared Euclidean distance between two embeddings satisfies $\|\mathbf{v}_i-\mathbf{v}_j\|_2^2=2-2\operatorname{sim}(\mathbf{v}_i,\mathbf{v}_j)$.

\subsection{Semantic Batching and Selection}
\label{subsec:batching}

\noindent\textbf{Semantic Batching.}
A batch should group questions that are semantically close, so that one shared reference bank is relevant to all of them and the answers generated within the batch are correlated enough to inform one another. We therefore form batches by clustering the target questions in semantic space. For notational simplicity and experimental convenience, assume $C\mid N$, so that $K=N/C$ and each batch contains $C$ questions. Given the embeddings $\mathbf{V} = \{\mathbf{v}_1, \dots, \mathbf{v}_N\}$ of the target questions, let $z_{ik} \in \{0,1\}$ indicate whether question $q_i$ is assigned to batch $B_k$. We then obtain the partition $\mathcal{B}$ by minimizing the intra-batch variance:
\begin{equation}
\label{eq:clustering}
\min_{\mathbf{Z},\,\{\mathbf{c}_k\}_{k=1}^{K}} \sum_{k=1}^K \sum_{i=1}^N z_{ik} \|\mathbf{v}_i - \mathbf{c}_k\|_2^2
\quad \text{s.t.} \quad
\textstyle\sum_{i=1}^N z_{ik} = C \;\; \forall k, \qquad
\textstyle\sum_{k=1}^K z_{ik} = 1 \;\; \forall i,
\end{equation}
where $\mathbf{c}_k = \frac{1}{C}\sum_{i=1}^N z_{ik}\mathbf{v}_i$ is the semantic centroid of batch $B_k$.

\noindent\textbf{Reference Bank Selection.}
For each batch $B_k$, we select the shared reference bank $S_k$ of Eq.~(\ref{eq:batched}) from the reference pool, preserving target-specific evidence before sharing it across the batch. Each target question $q_i \in B_k$ first retrieves its own top-$n$ references $R_i^{(n)} = \mathop{\arg\operatorname{Top}_{n}}_{s \in \mathcal{Q}_{pool}} \operatorname{sim}(\mathbf{v}_i, \mathbf{v}_s)$, where $n$ is the per-question budget and $\operatorname{sim}(\cdot,\cdot)$ is cosine similarity. The bank is then initialized as their union,
\begin{equation}
\label{eq:reference_union}
S_k^{\mathrm{init}} = \bigcup_{q_i \in B_k} R_i^{(n)},
\end{equation}
so that each question contributes its most relevant evidence while all questions in the batch have access to the same bank. The total budget is $m=nC$. If $\operatorname{card}(S_k^{\mathrm{init}})<m$ because target-specific references overlap, we fill the remaining slots by repeatedly adding the unselected example closest to the batch centroid, $s^* = \arg\max_{s \in \mathcal{Q}_{pool} \setminus S_k} \operatorname{sim}(\mathbf{v}_s,\mathbf{c}_k)$, until $\operatorname{card}(S_k)=m$ or the pool is exhausted. The resulting bank combines target-specific evidence with references representative of the overall batch semantics. To facilitate reference utilization, we annotate each target question with its relevant examples from the shared bank: for $q_i \in B_k$ we compute a wider retrieval set $R_i=\mathop{\arg\operatorname{Top}_{r}}_{s\in\mathcal{Q}_{pool}} \operatorname{sim}(\mathbf{v}_i,\mathbf{v}_s)$ with $r \ge n$ and take $\mathcal{A}_i = S_k \cap R_i$. The prompt presents the full bank $S_k$ once, while each target question is accompanied by its specific subset $\mathcal{A}_i$. This allows a question to leverage references introduced by other semantically related questions in the same batch without increasing the overall bank size.

\subsection{Curriculum-based Ordering}
\label{subsec:ordering}

\noindent\textbf{Target Question Ordering.}
Since each answer is generated conditioned only on what precedes it, we order questions within each batch $B_k$ so that answers produced earlier are the ones best able to inform those produced later. Let $\boldsymbol{\pi}=(\pi_1,\dots,\pi_C)$ denote a permutation of the question indices in $B_k$, and let $\mathcal{E}(q_i, S_k) = \max_{s \in S_k} \operatorname{sim}(\mathbf{v}_i, \mathbf{v}_s)$ be a heuristic easiness score, since similar references provide more relevant information and make a question easier to answer. Inspired by the curriculum learning paradigm, we apply a training-free heuristic that orders the questions from easy to hard:
\begin{equation}
\label{eq:argsort}
\boldsymbol{\pi}^{\mathrm{heur}} = \text{argsort}_{i \in B_k} \big(-\mathcal{E}(q_i, S_k)\big),
\end{equation}
which ensures $\mathcal{E}(q_{\pi^{\mathrm{heur}}_1}, S_k) \ge \dots \ge \mathcal{E}(q_{\pi^{\mathrm{heur}}_{C}}, S_k)$.

\noindent\textbf{Reference Bank Ordering.}
Given the ordered target sequence, we order the references within the shared bank by maximizing their semantic alignment with the target questions:
\begin{equation}
\label{eq:reference_order}
\boldsymbol{\phi}^* =
\arg\max_{\boldsymbol{\phi}}
\sum_{\ell=1}^{|S_k|}
\operatorname{sim}\left(\mathbf{v}_{s_{\phi(\ell)}}, \mathbf{v}_{q_{\pi^{\mathrm{heur}}_{g(\ell)}}}\right),
\end{equation}
where $\boldsymbol{\phi}$ is a permutation over references in $S_k$, and $g(\ell)=\min(C,\lceil \ell/n\rceil)$ maps the $\ell$-th reference slot to the corresponding target question. This places the references most relevant to the earlier target questions first, so that every question is preceded by the evidence it needs. Together, $\boldsymbol{\pi}^{\mathrm{heur}}$ and $\boldsymbol{\phi}^*$ instantiate the orderings of Eq.~(\ref{eq:batched}). Appendix~\ref{subsec:appendix_batch_prompt} shows how the resulting batches, reference banks, and orderings are serialized into the batch prediction prompt.

\section{Experiments}
\label{sec:experiments}

In this section, we conduct extensive experiments on four large-scale survey datasets across four LLMs to evaluate \model\ against the non-batched baseline. 
\begin{itemize}[
    label={},
    leftmargin=0pt,
    labelsep=0pt,
    itemindent=0pt,
    itemsep=0pt,
    topsep=2pt
]
    \item \hyperref[sec:rq1_effectiveness]{\textbf{(RQ1)}} How effective is \model\ compared with the non-batched baseline?
    \item \hyperref[sec:rq2_efficiency]{\textbf{(RQ2)}} How efficient is \model\ compared with the non-batched baseline?
    \item \hyperref[sec:batchsize]{\textbf{(RQ3)}} How does batch size affect the accuracy and cost of \model?
    \item \hyperref[sec:ablation]{\textbf{(RQ4)}} How do components of \model\ affect the performance?
\end{itemize}

\subsection{Experimental Setup}
\label{subsec:exp_setup}
\noindent\textbf{Datasets.}
We evaluate our framework on four public opinion survey datasets. World Values Survey (WVS) captures cross-national values and social attitudes; General Social Survey (GSS) measures long-term social trends in the United States; American National Election Studies (ANES) focuses on political behavior and electoral attitudes; and British Social Attitudes Survey (BSA) examines attitudes in the United Kingdom. We use the most recent wave of each dataset. Dataset filtering, respondent-profile construction, and reference--target pool construction are detailed in Appendix~\ref{sec:appendix_dataset_processing}.

\noindent\textbf{Large Language Models.}
We evaluate four widely used LLMs: DeepSeek-V4-Flash, DeepSeek-V4-Pro, Qwen3.7-Max, and GPT-4.1. These models span different capacity levels, allowing us to examine whether our framework consistently improves prediction performance across varying model sizes. For convenience, we use tiktoken to obtain a unified token count across all four models.

\noindent\textbf{Evaluation Metrics.}
We evaluate effectiveness using ACC, MAE, and F1, and efficiency using TPQ and SPQ. ACC measures prediction accuracy, while MAE measures the absolute difference between predicted and ground-truth option indices. F1 is computed separately for each question over its question-specific answer categories and averaged using question sample counts as weights. TPQ and SPQ denote the average prompt tokens and inference time per target question, respectively. Better performance corresponds to higher ACC and F1 and lower MAE, TPQ, and SPQ.

\noindent\textbf{Evaluation Setting.}
The non-batched baseline (w/o) follows the one-question-per-prompt paradigm and corresponds to the degenerate case of \model\ with batch size $C=1$. Each target question is predicted in its own prompt together with the respondent profile and its retrieved reference examples. We use $x$-budget to denote a setting with per-question reference budget $n=x$ (Section~\ref{subsec:batching}), i.e., each target question contributes up to $x$ retrieved references. All reported comparisons between the non-batched baseline (w/o) and SCBO (w) use the same reference budget to ensure a fair comparison.

\begin{table}[!t]
\definecolor{naturegray}{RGB}{140,140,140}
\newcommand{\deltapos}[1]{\textcolor{AweAIgreen}{\textbf{#1}}}
\newcommand{\deltaneg}[1]{\textcolor{naturegray}{\textbf{#1}}}
\caption{Effectiveness across reference budgets, LLMs, and datasets. $(w/o)$ denotes the non-batched baseline, $(w)$ denotes SCBO, and $\Delta$(\%) reports relative change. \textcolor{AweAIgreen}{\textbf{Green}} indicates improvement, while \textcolor{naturegray}{\textbf{gray}} indicates degradation.}
\vspace{-1.5em}
\label{tab:validity-results}
\begin{center}
\tiny
\setlength{\tabcolsep}{3pt}
\renewcommand{\arraystretch}{1.08}
\setlength{\aboverulesep}{0.2pt}
\setlength{\belowrulesep}{0.4pt}
\newcommand{\spacedhline}{\hline\noalign{\vskip 1.2pt}}
\newcommand{\spacedcdashline}{\noalign{\vskip 1.0pt}\cdashline{2-15}\noalign{\vskip 1.0pt}}
\resizebox{\textwidth}{!}{%
\begin{tabular}{lll*{3}{c}@{\hspace{4pt}}*{3}{c}@{\hspace{4pt}}*{3}{c}@{\hspace{4pt}}*{3}{c}}
\spacedhline
\multirow{2}{*}{\bf LLM} &
\multirow{2}{*}{\bf Budget} &
\multirow{2}{*}{\bf Setting} &
\multicolumn{3}{c}{\bf WVS} &
\multicolumn{3}{c}{\bf GSS} &
\multicolumn{3}{c}{\bf ANES} &
\multicolumn{3}{c}{\bf BSA} \\
\cmidrule(lr){4-6}\cmidrule(lr){7-9}\cmidrule(lr){10-12}\cmidrule(lr){13-15}\noalign{\vskip 0.2pt}
& & & ACC & MAE & F1 & ACC & MAE & F1 & ACC & MAE & F1 & ACC & MAE & F1 \\
\spacedhline
\multirow{9}{*}{\begin{tabular}{@{}c@{}}Deepseek\\V4-Flash\end{tabular}} & \multirow{3}{*}{1-budget} & w/o & 0.506 & 0.928 & 0.493 & 0.491 & 0.742 & 0.478 & 0.508 & 0.917 & 0.492 & 0.486 & 0.713 & 0.476 \\
 &  & w & 0.543 & 0.798 & 0.516 & 0.517 & 0.678 & 0.488 & 0.531 & 0.861 & 0.510 & 0.545 & 0.607 & 0.533 \\
 &  & \textbf{$\Delta$(\%)} & \deltapos{7.3} & \deltapos{14.0} & \deltapos{4.6} & \deltapos{5.3} & \deltapos{8.6} & \deltapos{2.1} & \deltapos{4.5} & \deltapos{6.1} & \deltapos{3.7} & \deltapos{12.1} & \deltapos{14.9} & \deltapos{12.0} \\
\spacedcdashline
 & \multirow{3}{*}{2-budget} & w/o & 0.522 & 0.883 & 0.512 & 0.528 & 0.676 & 0.510 & 0.527 & 0.868 & 0.508 & 0.492 & 0.674 & 0.485 \\
 &  & w & 0.548 & 0.761 & 0.525 & 0.547 & 0.615 & 0.527 & 0.556 & 0.793 & 0.535 & 0.552 & 0.589 & 0.532 \\
 &  & \textbf{$\Delta$(\%)} & \deltapos{5.0} & \deltapos{13.8} & \deltapos{2.6} & \deltapos{3.6} & \deltapos{9.0} & \deltapos{3.3} & \deltapos{5.5} & \deltapos{8.6} & \deltapos{5.3} & \deltapos{12.2} & \deltapos{12.6} & \deltapos{9.8} \\
\spacedcdashline
 & \multirow{3}{*}{3-budget} & w/o & 0.526 & 0.877 & 0.508 & 0.553 & 0.647 & 0.547 & 0.541 & 0.846 & 0.524 & 0.497 & 0.670 & 0.490 \\
 &  & w & 0.553 & 0.798 & 0.534 & 0.537 & 0.622 & 0.527 & 0.569 & 0.777 & 0.553 & 0.535 & 0.600 & 0.512 \\
 &  & \textbf{$\Delta$(\%)} & \deltapos{5.1} & \deltapos{9.0} & \deltapos{5.1} & \deltaneg{2.8} & \deltapos{3.9} & \deltaneg{3.7} & \deltapos{5.2} & \deltapos{8.2} & \deltapos{5.5} & \deltapos{7.6} & \deltapos{10.4} & \deltapos{4.6} \\
\spacedhline
\multirow{9}{*}{\begin{tabular}{@{}c@{}}Deepseek\\V4-Pro\end{tabular}} & \multirow{3}{*}{1-budget} & w/o & 0.505 & 0.868 & 0.487 & 0.495 & 0.696 & 0.480 & 0.514 & 0.884 & 0.504 & 0.501 & 0.677 & 0.484 \\
 &  & w & 0.551 & 0.749 & 0.524 & 0.532 & 0.642 & 0.512 & 0.534 & 0.836 & 0.513 & 0.542 & 0.590 & 0.526 \\
 &  & \textbf{$\Delta$(\%)} & \deltapos{9.1} & \deltapos{13.7} & \deltapos{7.6} & \deltapos{7.5} & \deltapos{7.8} & \deltapos{6.6} & \deltapos{3.9} & \deltapos{5.4} & \deltapos{1.7} & \deltapos{8.2} & \deltapos{12.9} & \deltapos{8.6} \\
\spacedcdashline
 & \multirow{3}{*}{2-budget} & w/o & 0.520 & 0.813 & 0.516 & 0.542 & 0.619 & 0.518 & 0.535 & 0.838 & 0.522 & 0.495 & 0.676 & 0.485 \\
 &  & w & 0.559 & 0.764 & 0.544 & 0.547 & 0.595 & 0.530 & 0.561 & 0.780 & 0.547 & 0.539 & 0.598 & 0.525 \\
 &  & \textbf{$\Delta$(\%)} & \deltapos{7.5} & \deltapos{6.0} & \deltapos{5.5} & \deltapos{0.9} & \deltapos{3.9} & \deltapos{2.2} & \deltapos{4.9} & \deltapos{6.9} & \deltapos{4.9} & \deltapos{8.9} & \deltapos{11.5} & \deltapos{8.1} \\
\spacedcdashline
 & \multirow{3}{*}{3-budget} & w/o & 0.533 & 0.817 & 0.518 & 0.552 & 0.622 & 0.541 & 0.556 & 0.792 & 0.548 & 0.498 & 0.707 & 0.482 \\
 &  & w & 0.567 & 0.746 & 0.552 & 0.569 & 0.581 & 0.559 & 0.586 & 0.731 & 0.569 & 0.537 & 0.588 & 0.524 \\
 &  & \textbf{$\Delta$(\%)} & \deltapos{6.4} & \deltapos{8.7} & \deltapos{6.7} & \deltapos{3.1} & \deltapos{6.6} & \deltapos{3.3} & \deltapos{5.4} & \deltapos{7.7} & \deltapos{3.9} & \deltapos{7.8} & \deltapos{16.8} & \deltapos{8.5} \\
\spacedhline
\multirow{9}{*}{\begin{tabular}{@{}c@{}}Qwen3.7\\Max\end{tabular}} & \multirow{3}{*}{1-budget} & w/o & 0.509 & 0.845 & 0.479 & 0.509 & 0.681 & 0.487 & 0.506 & 0.888 & 0.487 & 0.500 & 0.644 & 0.477 \\
 &  & w & 0.544 & 0.749 & 0.512 & 0.533 & 0.623 & 0.498 & 0.524 & 0.843 & 0.502 & 0.543 & 0.576 & 0.521 \\
 &  & \textbf{$\Delta$(\%)} & \deltapos{6.9} & \deltapos{11.4} & \deltapos{7.1} & \deltapos{4.7} & \deltapos{8.5} & \deltapos{2.3} & \deltapos{3.6} & \deltapos{5.1} & \deltapos{3.1} & \deltapos{8.6} & \deltapos{10.6} & \deltapos{9.4} \\
\spacedcdashline
 & \multirow{3}{*}{2-budget} & w/o & 0.520 & 0.837 & 0.502 & 0.558 & 0.602 & 0.533 & 0.532 & 0.819 & 0.515 & 0.501 & 0.652 & 0.485 \\
 &  & w & 0.554 & 0.738 & 0.532 & 0.552 & 0.593 & 0.521 & 0.546 & 0.798 & 0.529 & 0.550 & 0.586 & 0.516 \\
 &  & \textbf{$\Delta$(\%)} & \deltapos{6.5} & \deltapos{11.8} & \deltapos{6.1} & \deltaneg{1.1} & \deltapos{1.5} & \deltaneg{2.3} & \deltapos{2.6} & \deltapos{2.6} & \deltapos{2.9} & \deltapos{9.8} & \deltapos{10.1} & \deltapos{6.2} \\
\spacedcdashline
 & \multirow{3}{*}{3-budget} & w/o & 0.515 & 0.849 & 0.493 & 0.548 & 0.614 & 0.531 & 0.567 & 0.770 & 0.551 & 0.518 & 0.643 & 0.504 \\
 &  & w & 0.554 & 0.743 & 0.540 & 0.572 & 0.570 & 0.548 & 0.577 & 0.756 & 0.562 & 0.560 & 0.557 & 0.532 \\
 &  & \textbf{$\Delta$(\%)} & \deltapos{7.6} & \deltapos{12.5} & \deltapos{9.6} & \deltapos{4.4} & \deltapos{7.2} & \deltapos{3.1} & \deltapos{1.8} & \deltapos{1.8} & \deltapos{2.0} & \deltapos{8.1} & \deltapos{13.4} & \deltapos{5.6} \\
\spacedhline
\multirow{9}{*}{\begin{tabular}{@{}c@{}}GPT-4.1\end{tabular}} & \multirow{3}{*}{1-budget} & w/o & 0.491 & 0.947 & 0.480 & 0.512 & 0.701 & 0.500 & 0.505 & 0.906 & 0.489 & 0.510 & 0.668 & 0.495 \\
 &  & w & 0.530 & 0.812 & 0.505 & 0.531 & 0.637 & 0.512 & 0.525 & 0.840 & 0.512 & 0.542 & 0.618 & 0.519 \\
 &  & \textbf{$\Delta$(\%)} & \deltapos{7.9} & \deltapos{14.3} & \deltapos{5.1} & \deltapos{3.7} & \deltapos{9.1} & \deltapos{2.3} & \deltapos{4.0} & \deltapos{7.3} & \deltapos{4.8} & \deltapos{6.3} & \deltapos{7.5} & \deltapos{4.9} \\
\spacedcdashline
 & \multirow{3}{*}{2-budget} & w/o & 0.523 & 0.877 & 0.511 & 0.518 & 0.704 & 0.508 & 0.524 & 0.860 & 0.508 & 0.512 & 0.657 & 0.490 \\
 &  & w & 0.527 & 0.838 & 0.509 & 0.558 & 0.588 & 0.542 & 0.551 & 0.810 & 0.538 & 0.568 & 0.578 & 0.553 \\
 &  & \textbf{$\Delta$(\%)} & \deltapos{0.8} & \deltapos{4.4} & \deltaneg{0.3} & \deltapos{7.7} & \deltapos{16.5} & \deltapos{6.8} & \deltapos{5.2} & \deltapos{5.8} & \deltapos{5.9} & \deltapos{10.9} & \deltapos{12.0} & \deltapos{12.9} \\
\spacedcdashline
 & \multirow{3}{*}{3-budget} & w/o & 0.519 & 0.893 & 0.505 & 0.533 & 0.680 & 0.525 & 0.548 & 0.826 & 0.534 & 0.542 & 0.640 & 0.517 \\
 &  & w & 0.543 & 0.836 & 0.528 & 0.556 & 0.597 & 0.543 & 0.584 & 0.747 & 0.571 & 0.560 & 0.567 & 0.533 \\
 &  & \textbf{$\Delta$(\%)} & \deltapos{4.6} & \deltapos{6.4} & \deltapos{4.7} & \deltapos{4.3} & \deltapos{12.2} & \deltapos{3.5} & \deltapos{6.6} & \deltapos{9.6} & \deltapos{6.9} & \deltapos{3.3} & \deltapos{11.4} & \deltapos{3.1} \\
\hline
\end{tabular}%
}
\vspace{-2em}
\end{center}
\end{table}

\subsection{Implementation Details}
\label{subsec:implementation_details}

We use GPT-5.5 to extract semantics from raw survey questions, generating instantiated representations that filter out template noise. We embed the instantiated target and reference questions using \texttt{text-embedding-3-small}; reference answers are used as labels and excluded from the embedding input. These embeddings support clustering, retrieval, and ordering operations. Since the non-batched baseline and SCBO share the same instantiation and embedding process, the token costs are excluded from reported efficiency metrics to ensure a fair comparison.
For each dataset and reference budget, we select respondents with sufficient question coverage to support batch-size comparisons. Each respondent is assigned a fixed number of target questions, ensuring that under a $k$-budget setting, sufficient reference responses are available for reference selection. While our framework supports arbitrary batch sizes, we focus on divisor-based batch sizes for convenience, as they allow even partitioning without incomplete tail batches.
For each dataset--budget setting, we randomly split respondents into 15\% validation and 85\% test sets. We select the batch size by validation ACC for each dataset--model--budget configuration and report results on the test set.
Questions are grouped into fixed-size batches using KMeans++ clustering~\citep{arthur2007k} over instantiated question embeddings, followed by capacity-constrained assignment with the Hungarian algorithm~\citep{kuhn1955hungarian}. The complete capacity-constrained batching procedure is detailed in Appendix~\ref{sec:appendix_implementation}. Within each batch, we set $r=10$ for the wider target-specific retrieval used for relevant-reference annotation.

\subsection{Effectiveness of SCBO (RQ1)}
\label{sec:rq1_effectiveness}

Table~\ref{tab:validity-results} presents the overall performance of our framework across four datasets, four LLMs, and three reference budgets. Although a few configurations show small declines in ACC or F1, our method (w) generally achieves higher ACC, lower MAE, and higher F1 scores than the non-batched baseline (w/o). Across the 48 dataset--model--budget configurations, SCBO reduces MAE in all configurations, improves ACC in 46, and improves F1 in 45. On WVS, ACC gains reach at least 5\% in most configurations, while MAE reductions are observed across all reported configurations. The improvement is particularly notable on BSA, where ACC gains reach up to 12.2\% under the 1-budget and 2-budget settings. Similar positive trends are observed across most configurations on GSS and ANES. Moreover, our method improves prediction performance across models with different capability levels, including more capable models such as DeepSeek-V4-Pro and less capable models such as DeepSeek-V4-Flash. This generally positive pattern across diverse survey domains and different LLMs provides evidence of the broad generalizability of our approach, further supporting the effectiveness of our semantic batching and curriculum-based ordering strategies in improving LLM-based social survey prediction performance.

\begin{table}[!t]
\definecolor{natureblue}{RGB}{106,174,214}
\definecolor{naturegray}{RGB}{140,140,140}
\newcommand{\speedup}[1]{\textcolor{AweAIgreen}{\textbf{#1}}}
\caption{Efficiency across reference budgets, LLMs, and datasets. Columns 1--3 denote the per-question reference budget; $(w/o)$ denotes the non-batched baseline and $(w)$ denotes SCBO. \textcolor{AweAIgreen}{\textbf{Green}} indicates improvement, while \textcolor{naturegray}{\textbf{gray}} indicates degradation.}
\vspace{-1.5em}

\label{tab:efficiency-results}
\begin{center}
\tiny
\setlength{\tabcolsep}{3pt}
\renewcommand{\arraystretch}{1.08}
\setlength{\aboverulesep}{0.2pt}
\setlength{\belowrulesep}{0.4pt}
\newcommand{\spacedhline}{\hline\noalign{\vskip 1.2pt}}
\newcommand{\spacedcdashline}{\noalign{\vskip 1.0pt}\cdashline{2-15}\noalign{\vskip 1.0pt}}
\resizebox{\textwidth}{!}{%
\begin{tabular}{lll*{3}{c}@{\hspace{4pt}}*{3}{c}@{\hspace{4pt}}*{3}{c}@{\hspace{4pt}}*{3}{c}}
\spacedhline
\multirow{2}{*}{\bf LLM} &
\multirow{2}{*}{\bf Metric} &
\multirow{2}{*}{\bf Setting} &
\multicolumn{3}{c}{\bf WVS} &
\multicolumn{3}{c}{\bf GSS} &
\multicolumn{3}{c}{\bf ANES} &
\multicolumn{3}{c}{\bf BSA} \\
\cmidrule(lr){4-6}\cmidrule(lr){7-9}\cmidrule(lr){10-12}\cmidrule(lr){13-15}\noalign{\vskip 0.2pt}
& & & 1 & 2 & 3 & 1 & 2 & 3 & 1 & 2 & 3 & 1 & 2 & 3 \\
\spacedhline
\multirow{6}{*}{\begin{tabular}{@{}c@{}}DeepSeek\\V4-Flash\end{tabular}}
& \multirow{3}{*}{TPQ} & w/o & 577 & 661 & 747 & 589 & 659 & 727 & 621 & 698 & 774 & 541 & 613 & 681 \\
& & w & 194 & 280 & 380 & 170 & 253 & 401 & 181 & 257 & 344 & 187 & 247 & 329 \\
& & Redu.(\%) & \speedup{66.3} & \speedup{57.7} & \speedup{49.1} & \speedup{71.1} & \speedup{61.6} & \speedup{44.8} & \speedup{70.8} & \speedup{63.3} & \speedup{55.5} & \speedup{65.4} & \speedup{59.7} & \speedup{51.7} \\
\spacedcdashline
& \multirow{3}{*}{SPQ} & w/o & 0.947 & 1.085 & 0.839 & 0.851 & 0.927 & 0.829 & 0.889 & 1.026 & 0.859 & 1.162 & 1.063 & 0.978 \\
& & w & 0.093 & 0.115 & 0.146 & 0.105 & 0.114 & 0.142 & 0.086 & 0.093 & 0.100 & 0.101 & 0.117 & 0.124 \\
& & Speedup & \speedup{10.2$\times$} & \speedup{9.4$\times$} & \speedup{5.8$\times$} & \speedup{8.2$\times$} & \speedup{8.1$\times$} & \speedup{5.8$\times$} & \speedup{10.3$\times$} & \speedup{11.0$\times$} & \speedup{8.6$\times$} & \speedup{11.6$\times$} & \speedup{9.1$\times$} & \speedup{7.9$\times$} \\
\spacedhline
\multirow{6}{*}{\begin{tabular}{@{}c@{}}DeepSeek\\V4-Pro\end{tabular}}
& \multirow{3}{*}{TPQ} & w/o & 577 & 661 & 747 & 589 & 659 & 727 & 621 & 698 & 774 & 541 & 613 & 681 \\
& & w & 194 & 280 & 363 & 170 & 278 & 325 & 181 & 257 & 327 & 173 & 268 & 329 \\
& & Redu.(\%) & \speedup{66.3} & \speedup{57.7} & \speedup{51.4} & \speedup{71.1} & \speedup{57.9} & \speedup{55.3} & \speedup{70.8} & \speedup{63.3} & \speedup{57.8} & \speedup{68.0} & \speedup{56.3} & \speedup{51.7} \\
\spacedcdashline
& \multirow{3}{*}{SPQ} & w/o & 1.148 & 1.435 & 1.423 & 1.241 & 1.434 & 1.634 & 1.130 & 1.233 & 1.454 & 1.349 & 1.346 & 1.353 \\
& & w & 0.137 & 0.169 & 0.175 & 0.146 & 0.157 & 0.179 & 0.143 & 0.153 & 0.160 & 0.152 & 0.170 & 0.204 \\
& & Speedup & \speedup{8.4$\times$} & \speedup{8.5$\times$} & \speedup{8.2$\times$} & \speedup{8.5$\times$} & \speedup{9.1$\times$} & \speedup{9.1$\times$} & \speedup{7.9$\times$} & \speedup{8.1$\times$} & \speedup{9.1$\times$} & \speedup{8.9$\times$} & \speedup{7.9$\times$} & \speedup{6.6$\times$} \\
\spacedhline
\multirow{6}{*}{\begin{tabular}{@{}c@{}}Qwen3.7\\Max\end{tabular}}
& \multirow{3}{*}{TPQ} & w/o & 577 & 661 & 747 & 589 & 659 & 727 & 621 & 698 & 774 & 541 & 613 & 681 \\
& & w & 210 & 287 & 446 & 170 & 242 & 316 & 181 & 465 & 541 & 193 & 247 & 320 \\
& & Redu.(\%) & \speedup{63.5} & \speedup{56.5} & \speedup{40.2} & \speedup{71.1} & \speedup{63.3} & \speedup{56.5} & \speedup{70.8} & \speedup{33.5} & \speedup{30.1} & \speedup{64.3} & \speedup{59.7} & \speedup{53.0} \\
\spacedcdashline
& \multirow{3}{*}{SPQ} & w/o & 0.899 & 0.985 & 1.173 & 1.346 & 1.671 & 1.155 & 1.326 & 1.197 & 2.960 & 1.271 & 1.331 & 1.748 \\
& & w & 0.275 & 0.286 & 0.301 & 0.288 & 0.298 & 0.315 & 0.275 & 0.280 & 0.283 & 0.279 & 0.309 & 0.319 \\
& & Speedup & \speedup{3.3$\times$} & \speedup{3.4$\times$} & \speedup{3.9$\times$} & \speedup{4.7$\times$} & \speedup{5.6$\times$} & \speedup{3.7$\times$} & \speedup{4.8$\times$} & \speedup{4.3$\times$} & \speedup{10.5$\times$} & \speedup{4.6$\times$} & \speedup{4.3$\times$} & \speedup{5.5$\times$} \\
\spacedhline
\multirow{6}{*}{GPT-4.1}
& \multirow{3}{*}{TPQ} & w/o & 577 & 661 & 747 & 589 & 659 & 727 & 621 & 698 & 774 & 541 & 613 & 681 \\
& & w & 194 & 280 & 371 & 170 & 253 & 325 & 200 & 465 & 429 & 227 & 252 & 329 \\
& & Redu.(\%) & \speedup{66.3} & \speedup{57.7} & \speedup{50.3} & \speedup{71.1} & \speedup{61.6} & \speedup{55.3} & \speedup{67.7} & \speedup{33.5} & \speedup{44.5} & \speedup{58.1} & \speedup{58.8} & \speedup{51.7} \\
\spacedcdashline
& \multirow{3}{*}{SPQ} & w/o & 1.663 & 1.724 & 1.738 & 1.979 & 1.686 & 1.669 & 1.674 & 1.752 & 1.774 & 1.590 & 1.617 & 1.713 \\
& & w & 0.174 & 0.170 & 0.188 & 0.141 & 0.170 & 0.213 & 0.143 & 0.144 & 0.161 & 0.158 & 0.173 & 0.184 \\
& & Speedup & \speedup{9.6$\times$} & \speedup{10.1$\times$} & \speedup{9.3$\times$} & \speedup{14.0$\times$} & \speedup{9.9$\times$} & \speedup{7.9$\times$} & \speedup{11.7$\times$} & \speedup{12.1$\times$} & \speedup{11.0$\times$} & \speedup{10.1$\times$} & \speedup{9.4$\times$} & \speedup{9.3$\times$} \\
\hline
\end{tabular}%
}
\vspace{-1em}
\end{center}
\end{table}

\subsection{Efficiency of SCBO (RQ2)}
\label{sec:rq2_efficiency}


Table~\ref{tab:efficiency-results} compares the token cost and inference time of the non-batched baseline and our framework. SCBO consistently reduces both TPQ and SPQ across all evaluated configurations. For token cost, which is our primary optimization objective, the reduction ranges from 30.1\% to 71.1\%, with an average reduction of approximately 58.0\%; reductions exceed 50\% in most configurations. Although SCBO primarily targets token efficiency, batching also substantially reduces inference time by amortizing communication latency and other per-request overheads across multiple questions. The observed speedups range from 5.8$\times$ to 11.6$\times$ for DeepSeek-V4-Flash, 6.6$\times$ to 9.1$\times$ for DeepSeek-V4-Pro, 7.9$\times$ to 14.0$\times$ for GPT-4.1, and 3.3$\times$ to 10.5$\times$ for Qwen3.7-Max. These results indicate that SCBO can improve the efficiency of large-scale LLM-based social survey prediction in terms of both token consumption and inference time in realistic large-scale social survey deployment scenarios.

\begin{figure}[!t]
    \centering
    \includegraphics[width=\textwidth]{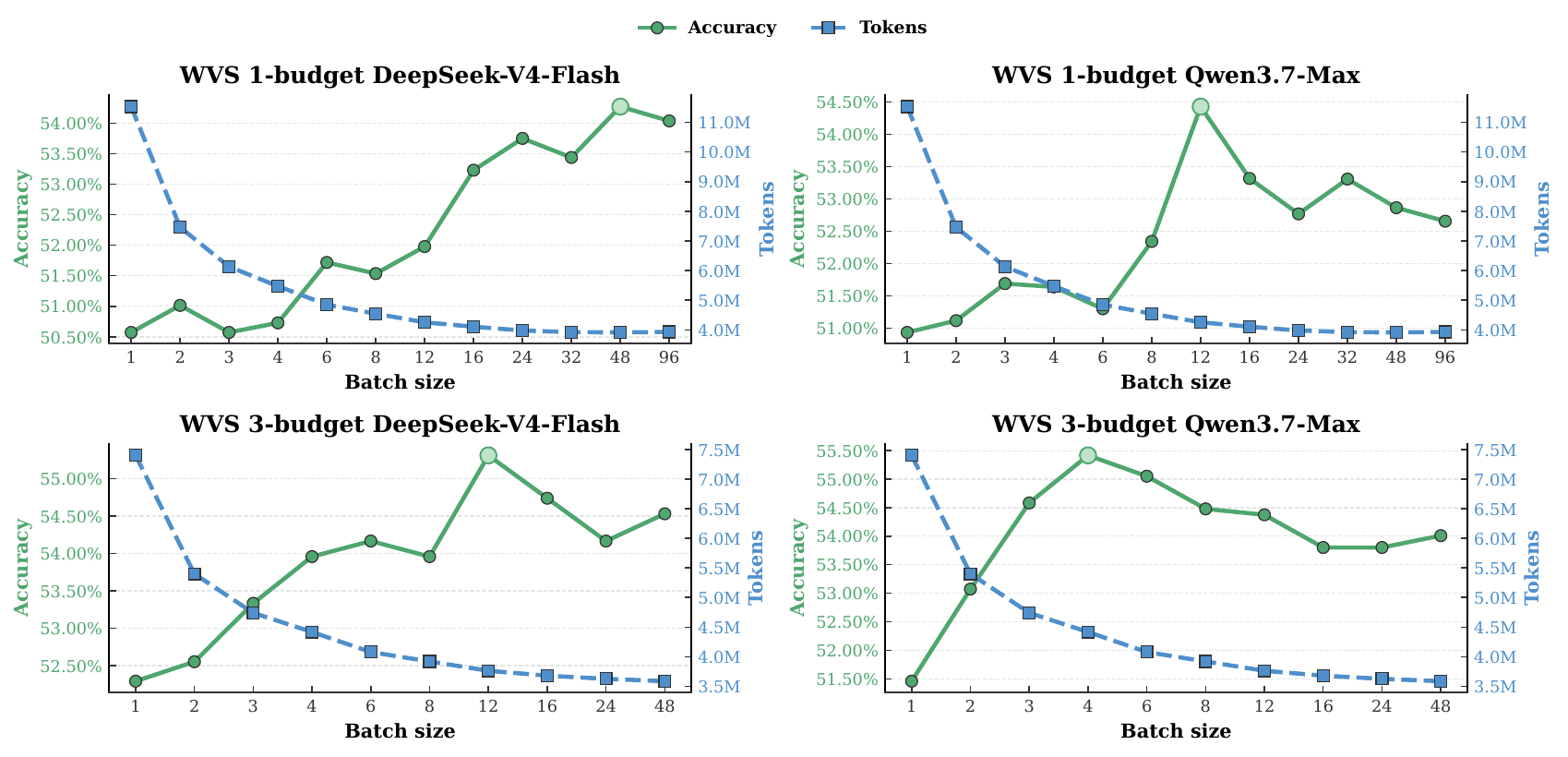}
    \vspace{-1em}
    \caption{Accuracy and token cost across batch sizes on WVS. The figure compares 1-budget and 3-budget settings using DeepSeek-V4-Flash and Qwen3.7-Max.}
    \vspace{-1em}
    \label{fig:wvs-batch-size-acc-token}
\end{figure}

\subsection{Impact of Batch Size (RQ3)}
\label{sec:batchsize}

Figure~\ref{fig:wvs-batch-size-acc-token} illustrates how batch size affects ACC and TPQ on WVS under the 1-budget and 3-budget settings with DeepSeek-V4-Flash and Qwen3.7-Max. Our framework demonstrates strong robustness to batch size selection: across all tested batch sizes, the batched method consistently outperforms the non-batched baseline in both accuracy and token cost. As batch size increases, TPQ decreases monotonically due to greater context amortization, achieving substantial token reductions even at non-optimal batch sizes. ACC exhibits a non-monotonic pattern, initially improving with moderate batch sizes (around 8–16 for DeepSeek-V4-Flash), then plateauing or slightly declining at very large batch sizes, suggesting that excessively large batches may introduce attentional disruption despite semantic clustering. These findings confirm that our framework reliably improves over single-query prompting regardless of batch size choice, while also highlighting that moderate batch sizes offer the best balance between token efficiency and prediction accuracy.

\begin{figure}[!t]
    \centering
    \includegraphics[width=\textwidth]{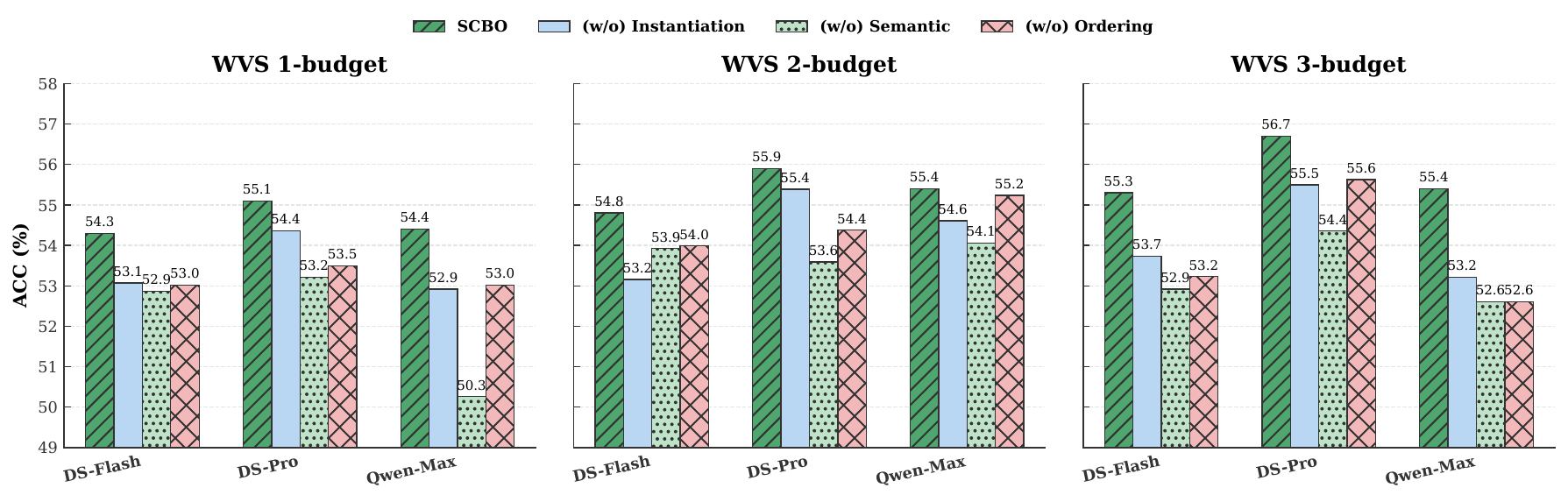}
    \vspace{-1em}
    \caption{Ablation study on WVS across different component variants and reference budgets.}
    \vspace{-1em}
    \label{fig:wvs-ablation-acc}
\end{figure}

\subsection{Ablation Study (RQ4)}
\label{sec:ablation}

In this section, we perform ablation studies to examine the contribution of each component in SCBO. All experiments are conducted on WVS across three LLMs (DeepSeek-V4-Flash, DeepSeek-V4-Pro, and Qwen3.7-Max) under 1-budget, 2-budget, and 3-budget. The following variants are evaluated:
\noindent\textcolor{AblationBlue}{\textbf{(w/o) Instantiation}}: This variant excludes question instantiation. Raw survey questions are directly embedded without extracting core semantics, introducing template noise into clustering and retrieval.
\textcolor{AblationGreen}{\textbf{(w/o) Semantic}}: This variant removes semantic batching. Questions are randomly grouped into batches instead of being clustered by semantic similarity, which increases attentional disruption and reduces reference utility across batch questions.
\textcolor{AblationPink}{\textbf{(w/o) Ordering}}: This variant omits curriculum-based ordering. Both target questions and reference examples are arranged randomly instead of following the easy-to-hard curriculum and alignment.

Figure~\ref{fig:wvs-ablation-acc} reports accuracy for these variants across models and reference budgets. SCBO outperforms all ablation variants. Removing semantic batching (\textcolor{AblationGreen}{\textbf{(w/o) Semantic}}) results in the largest performance drop, confirming that semantic clustering is the most critical component: it makes the shared bank relevant to each question and groups mutually informative questions. Removing ordering (\textcolor{AblationPink}{\textbf{(w/o) Ordering}}) degrades performance, highlighting the importance of curriculum-based ordering for exploiting autoregressive dependencies. Removing question instantiation (\textcolor{AblationBlue}{\textbf{(w/o) Instantiation}}) shows a smaller decline, indicating that filtering template noise improves embedding quality. These results validate that both modules contribute synergistically, with semantic batching playing the dominant role, followed by ordering, while instantiation provides a useful basis for both.

\section{Conclusion}
\label{sec:conclusion}

In this paper, we addressed three limitations of the one-question-per-prompt paradigm in LLM-based social survey prediction: it re-encodes the same context for every question, restricts each target question to a narrow subset of the respondent's reference responses, and predicts every answer in isolation. We presented SCBO, which batches multiple questions into a single prompt, thereby amortizing the shared context, allowing target questions to share a broader pool of reference responses, and enabling correlated answers to inform one another. Operating on compact question representations produced by a question instantiation step that filters out template noise, SCBO integrates two modules: semantic batching, which groups similar questions and selects a shared reference bank balancing target-specific relevance with batch-level sharing; and curriculum-based ordering, which arranges questions and references so that earlier answers inform later ones. Extensive experiments across diverse datasets and LLMs demonstrate that SCBO generally outperforms the non-batched baseline, achieving superior accuracy while reducing token costs by over 50\% and accelerating inference by 3.3--14.0$\times$. Future work will explore model-adaptive batching, which constructs batches based on feedback from the LLM itself rather than relying on heuristic clustering strategies.

\section*{AI Use Statement}
AI models serve as experimental components in this work. GPT-5.5 is used for Question Instantiation, extracting compact topic, intent, and entity representations from raw survey items. The \texttt{text-embedding-3-small} model encodes these representations into semantic vectors used for clustering, reference retrieval, and ordering. The evaluated LLMs generate predicted survey responses under both the non-batched baseline and SCBO settings. We also used generative AI tools to assist with English translation and language editing. The research idea, theoretical formulation, methodology, experimental design, implementation decisions, and interpretation of results are the authors' own. All AI-assisted content was reviewed and verified by the authors, who take full responsibility for the final text, claims, code, figures, and reported results.

\section*{Ethics Statement}
This work conducts secondary analysis of publicly released survey datasets, including WVS, GSS, ANES, and BSA. We use de-identified respondent records in accordance with the data-use conditions and do not attempt to identify individual respondents. No new human participants were recruited, and results are reported only in aggregate. Because model-based survey prediction may inherit biases from survey data and LLMs, the predictions should not be treated as substitutes for real respondents or used for high-stakes decisions about individuals or demographic groups. The study is intended to evaluate the methodological feasibility and efficiency of LLM-based social survey prediction.

\section*{Reproducibility Statement}
We have made an effort to make SCBO reproducible. The complete framework is specified in Section~\ref{sec:methodology}, and the experimental setup and implementation details are provided in Section~\ref{sec:experiments}. Appendix~\ref{sec:appendix_implementation} describes the capacity-constrained batching procedure; Appendix~\ref{sec:appendix_dataset_processing} documents dataset filtering, respondent-profile construction, reference--target pool construction, and evaluation statistics; and Appendix~\ref{sec:appendix_prompts} provides the prompt templates used for question instantiation and batch prediction. Source code and supporting materials are available at \url{https://anonymous.4open.science/r/SCBO-41D8}.

\bibliography{iclr2027_conference}
\bibliographystyle{iclr2027_conference}

\appendix
\section{Related Work}
\label{sec:related_work}
This section reviews three related areas: LLM-based social simulation, LLM-based survey respondent simulation, and batch prompting.

\noindent\textbf{Large Language Models for Social Simulation.}
Social simulation seeks to explain how macro-level collective phenomena arise from micro-level individual interactions.
Early formal treatments model this process through mathematical frameworks in which localized update rules---including Friedkin--Johnsen opinion dynamics and bounded-confidence protocols---drive the evolution of agent states across network topologies~\citep{friedkin1990social, Deffuant2000MixingBA, Kan2023BCM}.
Agent-based modeling builds on this bottom-up perspective by populating structured environments with autonomous agents whose interactions give rise to emergent societal patterns~\citep{epstein1996growing, Watts1998Collective, Barabsi1999EmergenceOS, Bernstein2013agent}.
LLM-powered simulation has recently advanced this paradigm by replacing hand-coded behavioral rules with language-grounded agents capable of reasoning over rich demographic profiles and contextual information, substantially expanding the expressiveness of synthetic societies~\citep{park2023generative, yu2025researchtown}.
Within networked multi-agent frameworks, these LLM agents have been applied to investigate a range of collective phenomena, encompassing opinion shift dynamics, emotional contagion, and the spread of information or misinformation~\citep{gao2025s3socialnetworksimulationlarge, Ferraro2025echo, chuang2024simulating, liu2025fakenews}.
Beyond social networks, LLMs have been employed to simulate individual user behaviors in interactive systems and social governance~\citep{wang2025user, li2026towards}.
In this work, we focus on the individual level of this simulation stack: accurately predicting each respondent's answers to survey questions as a foundation for downstream collective modeling.

\noindent\textbf{Simulating Survey Respondents with LLMs.}
Traditional survey methods face challenges including high costs, logistical constraints, and limited scalability \citep{wright2010survey,heffetz2019difficulty,kalton2009methods}. Recent advances in Large Language Models have opened new possibilities for social simulation and survey research \citep{argyle2023out, park2023generative, aher2023using, cao2023assessing, zhou2025should, durmus2023towards,wang2025user, chen2026benchmarking}. Early works demonstrated that LLMs can exhibit human-like behaviors and attitudes when properly conditioned \citep{argyle2023out, park2023generative,zhou2024exploring}. Building on these findings, researchers have explored using LLMs to simulate survey respondents at scale \citep{santurkar2023whose, simmons2023moral, bisbee2024synthetic, cao2023assessing}.
These approaches condition LLMs on demographic profiles and observed reference responses to predict answers to survey questions. However, most existing work adopts a one-question-per-prompt paradigm, which incurs prohibitive token costs for large-scale simulations. Our work addresses this limitation by proposing an efficient batching and ordering framework that maintains prediction quality while reducing computational costs.

\noindent\textbf{Batch Prompting.}
Batch prompting improves LLM inference efficiency by processing multiple
instances in a single prompt and amortizing shared instructions and
demonstrations~\citep{cheng2023batch}. Cheng et al. also examine semantic and
diversity-based grouping, but find no consistent improvement over random
batching. Lin et al.~\citep{lin2024batchprompt} show that batched predictions
are sensitive to instance position and order, and propose Batch Permutation and
Ensembling together with early stopping to improve stability while controlling
additional cost. Cascaded batch prompting further separates reasoning from
output-symbol grounding to improve batched
classification~\citep{hoshino2026cascaded}.
Although closely related, existing batch-prompting methods differ from SCBO in
both objective and problem structure. They primarily batch independent
instances to reduce inference cost, whereas survey questions for the same
respondent are often semantically related and grounded in shared respondent
evidence. SCBO exploits this structure by constructing a broader
respondent-specific reference bank shared across questions and by ordering
questions from easy to hard with semantically aligned references. These designs
allow SCBO to improve prediction accuracy in addition to inference efficiency.

\section{Implementation Details}
\label{sec:appendix_implementation}

\begin{figure}[!t]
    \centering
    \includegraphics[width=\textwidth]{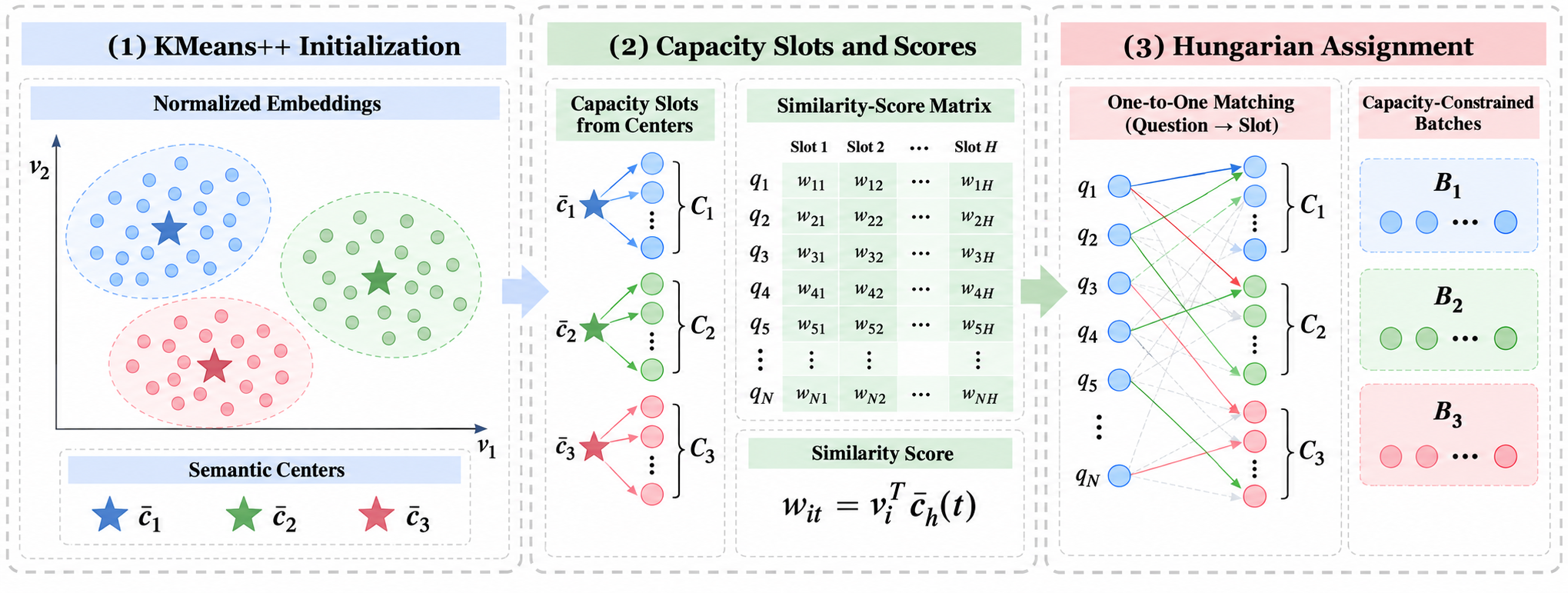}
    \vspace{-1em}
    \caption{Capacity-constrained semantic batching. KMeans++ identifies semantic centers from question embeddings. Each center is expanded into slots according to its capacity, and the Hungarian algorithm computes a one-to-one assignment that produces capacity-constrained batches.}
    \label{fig:capacity-constrained-batching}
    \vspace{-1em}
\end{figure}

For notational simplicity and to match the experimental settings, the main text assumes that the number of target questions is divisible by the batch capacity, so that every batch contains exactly $C$ questions; here, we consider the general non-divisible case, in which the final batch may contain fewer than $C$ questions (Appendix~\ref{subsec:appendix_notation}). We then describe how to obtain the resulting capacity-constrained partition in practice (Appendix~\ref{subsec:appendix_capacity_batching}), as illustrated in Figure~\ref{fig:capacity-constrained-batching}. For each respondent, we first use KMeans++ to obtain initial semantic centers and then apply the Hungarian algorithm to assign target questions to slots with prescribed capacities. The resulting assignment satisfies the prescribed capacities and provides a deterministic approximation to the fixed-size clustering objective, and all operations are performed independently for each respondent.

\subsection{Notation and Boundary Cases}
\label{subsec:appendix_notation}

For a respondent $u$, let $\mathcal{Q}^{u}_{target}$ contain $N_u$ target questions and let $\mathcal{Q}^{u}_{pool}$ contain $H_u$ observed question--answer pairs. We use $C$ as the maximum number of target questions in a batch, $n$ as the per-question reference budget, and $r$ as the retrieval depth used to expose relevant reference IDs, where $r\geq n$. The number of batches is
\begin{equation}
K_u=\left\lceil\frac{N_u}{C}\right\rceil.
\end{equation}
The first $K_u-1$ batches have capacity $C$, and the final batch has capacity $N_u-C(K_u-1)$. Thus, for a batch $B_k$ of size $C_k=|B_k|$, its shared reference-bank budget is respondent- and batch-specific:
\begin{equation}
m_k=nC_k.
\end{equation}

If $C\geq N_u$, then $K_u=1$ and all target questions form a single batch. If $N_u$ is not divisible by $C$, only the final batch has capacity smaller than $C$. When $H_u<m_k$, the reference bank cannot reach its nominal budget and contains at most $H_u$ reference pairs. These cases do not alter the batching assignment, but they affect the realized batch or reference-bank size.

\subsection{Capacity-Constrained Semantic Batching}
\label{subsec:appendix_capacity_batching}

We approximate the fixed-size clustering objective in Section~\ref{subsec:batching} in two stages. First, we run KMeans with KMeans++ initialization, $K_u$ clusters, ten initializations, and respondent-specific random seed $s_u=s+u$. The resulting cluster centers are then L2-normalized. Second, we enforce the prescribed batch capacities through a linear-assignment problem.

Specifically, let $\bar{\mathbf c}_1,\ldots,\bar{\mathbf c}_{K_u}$ be the normalized KMeans centers, and create $C_k$ identical assignment slots for cluster $k$. If slot $t$ belongs to cluster $h(t)$, its assignment score for target question $q_i$ is
\begin{equation}
w_{it}=\mathbf v_i^{\top}\bar{\mathbf c}_{h(t)}.
\end{equation}
We solve
\begin{equation}
\max_{x}\sum_{i=1}^{N_u}\sum_{t=1}^{N_u}x_{it}w_{it}
\quad\text{s.t.}\quad
\sum_t x_{it}=1,\quad
\sum_i x_{it}=1,\quad
x_{it}\in\{0,1\},
\end{equation}
using the Hungarian algorithm. Mapping each assigned slot to its associated cluster produces batches with the prescribed capacities. Given the respondent-specific seed, this provides a deterministic capacity-constrained approximation to Equation~\ref{eq:clustering}.

\begin{figure}[!t]
    \centering
    \includegraphics[width=\textwidth]{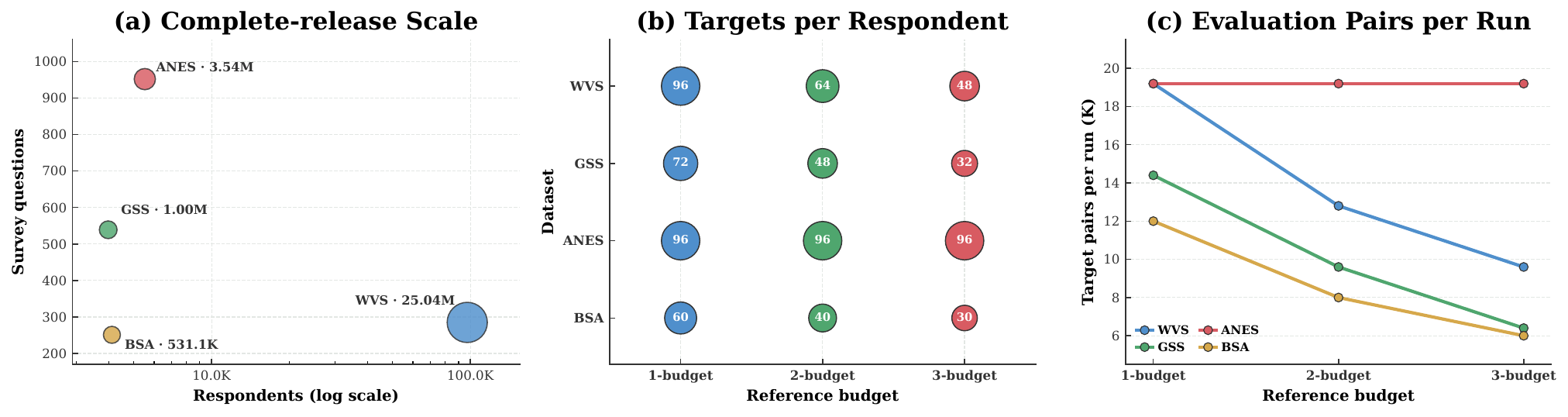}
    \vspace{-1em}
    \caption{Dataset scale and evaluation allocation. (a) Complete-release scale, where the horizontal axis denotes the number of respondents, the vertical axis denotes the number of survey questions, and marker area represents the number of observed respondent--question pairs. (b) Number of target questions assigned to each respondent under reference budgets. (c) Total target pairs evaluated in each method--model run. WVS uses Wave 7, while GSS, ANES, and BSA use their 2024 releases.}
    \label{fig:dataset-evaluation-overview}
    \vspace{-1em}
\end{figure}

\subsubsection{Inference-Time Measurement}
All models are accessed through OpenAI-compatible APIs: the official DeepSeek endpoint (\url{https://api.deepseek.com}) for DeepSeek-V4-Flash and DeepSeek-V4-Pro, Alibaba Cloud DashScope (\url{https://dashscope.aliyuncs.com/compatible-mode/v1}) for Qwen3.7-Max, and an OpenAI-compatible third-party gateway (\url{https://aiapi.world/v1}) for GPT-4.1. We use greedy decoding with temperature 0, disable streaming and model thinking, and set the maximum output length to 4,096 tokens. Both the non-batched baseline and SCBO use 30 concurrent workers. Failed or empty requests are retried up to five times with exponential backoff starting at two seconds.
For each request, elapsed time is measured with a monotonic wall-clock timer from prompt construction through API completion and output parsing. SPQ is computed by summing the de-duplicated request times and dividing by the number of target questions. It therefore measures aggregate request latency amortized per question rather than end-to-end wall-clock time under parallel execution. The client configuration is held fixed across methods, but we cannot control provider-side region, routing, network conditions, or server load; the reported speedups should be interpreted as measurements under our API environment rather than provider-independent latency guarantees.

\section{Dataset Processing}
\label{sec:appendix_dataset_processing}

We standardize the four survey releases into respondent--question prediction records and construct disjoint reference and target pools. All compared methods use the same processed records and respondent-specific pool assignments.

\subsection{Survey Sources and Response Processing}
\label{subsec:appendix_dataset_sources}

We use World Values Survey Wave 7 (WVS), General Social Survey 2024 (GSS), American National Election Studies 2024 (ANES), and British Social Attitudes 2024 (BSA). Figure~\ref{fig:dataset-evaluation-overview}(a) summarizes the complete-release scale underlying the exhaustive cost estimate in Table~\ref{tab:cost_estimation}. An observed pair comprises a respondent, a survey question, and the respondent's recorded answer.

Question text and response options are extracted from each survey's official codebook and matched to the corresponding respondent-level variables. We retain closed-ended, single-choice questions with complete option sets. Open-ended questions, multiple-selection questions, multi-field questions, and items containing unresolved template placeholders are excluded.

We remove administrative and non-substantive response categories, including missing, skipped, refused, invalid, and interviewer-only codes. Substantive choices, such as neutral positions and respondent-selectable ``other'' categories, are retained. Questions without a valid answer set after this filtering are discarded.

The remaining options are ordered according to their source codes and mapped to consecutive indices $\{0,\ldots,L_q-1\}$ for a question with $L_q$ valid choices. Each respondent answer is transformed using the corresponding question-specific mapping, and records that cannot be mapped to a retained option are removed. Consequently, every retained answer $a_{ui}$ satisfies
\begin{equation}
0\leq a_{ui}<L_{q_i}.
\end{equation}

\subsection{Respondent Profiles}
\label{subsec:appendix_persona_processing}

Background and demographic items are separated from prediction questions and converted into a third-person natural-language profile for each respondent. Profile items do not enter either the reference pool or the target pool, and respondents without an available profile are excluded.

\subsection{Reference--Target Pool Construction}
\label{subsec:appendix_pool_construction}

For each dataset, we collect all valid non-profile responses and construct respondent-specific reference and target pools separately for each reference budget. Under a given budget, all methods receive identical respondent profiles, references, and target questions.

Let $k\in\{1,2,3\}$ denote the per-target reference budget and $T_{d,k}$ the number of target questions assigned to each retained respondent in dataset $d$. Valid non-profile responses are randomly ordered using seed 42. The first $T_{d,k}$ records define the target pool, and the subsequent $kT_{d,k}$ records define the reference pool. The two pools therefore satisfy
\begin{equation}
\mathcal Q^{u}_{target}\cap\mathcal Q^{u}_{pool}=\emptyset,
\qquad
\operatorname{card}(\mathcal Q^{u}_{target})=T_{d,k},
\qquad
\operatorname{card}(\mathcal Q^{u}_{pool})=kT_{d,k}.
\end{equation}
A respondent is eligible only if at least $(k+1)T_{d,k}$ valid non-profile responses are available. Respondents with insufficient coverage are excluded, and target responses are never reused as references. Pool assignments remain unchanged across models, batch sizes, and compared methods.

The per-respondent target allocation is shown in Figure~\ref{fig:dataset-evaluation-overview}(b). Because the coverage requirement depends on $k$ and $T_{d,k}$, the eligible respondent set may differ across reference budgets.

SCBO retrieves references exclusively from the corresponding respondent's reference pool. Embeddings are computed from question text and answer options; respondent answers serve only as labels for retrieved references and are excluded from the embedding input.

\subsection{Evaluation Data Statistics}
\label{subsec:appendix_actual_data_usage}

For each dataset--budget setting, respondents are randomly split into 15\% validation and 85\% test sets. Batch size is selected by validation ACC for each dataset--model--budget configuration, and all reported metrics are computed on the shared test set. Figure~\ref{fig:dataset-evaluation-overview}(c) reports the number of test target pairs per method--model run.


\section{Prompt Templates}
\label{sec:appendix_prompts}

We report the exact templates used for question instantiation (Section~\ref{subsec:instantiation}) and batch prediction (Section~\ref{subsec:ordering}).

\subsection{Question Instantiation Prompt}
\label{subsec:appendix_instantiation_prompt}

The following prompt implements $\mathcal{P}_{extract}$ in Section~\ref{subsec:instantiation}. Given a raw question $q_i$, the instantiation model $\mathcal{M}$ produces $\tilde{q}_i=\{T_i,I_i,E_i\}$, representing its topic, core intent, and key entities. The returned \texttt{instance} field is embedded by $f_{emb}$. The placeholder \texttt{\{question\_payload\}} contains the question identifier, text, and answer options.

\begin{PromptBox}{Prompt A: Question Instantiation}
You extract compact structured instances from survey questions for semantic embedding.
Remove generic survey wording, respondent-addressing phrases, and repeated answer-scale phrasing.
Keep only the core topic, entities, and intent.
Return valid JSON only with these keys:
{ "topic": "...", "intent": "...", "entities": ["..."], "instance": "Topic: ...; Intent: ...; Entities: ..." }

Survey question:
{question_payload}

[The {question_payload} placeholder is replaced with:]
Question ID: {qid}
Question: {question_text}
Options:
{options_text}
\end{PromptBox}

\subsection{Batch Prediction Prompt}
\label{subsec:appendix_batch_prompt}

For each batch $B_k$, the target LLM receives the respondent profile, shared reference bank $S_k$, and ordered target questions through the following prompt.

\begin{PromptBox}{Prompt B: Batch Prediction}
You are predicting how a specific person would answer survey questions.

## Person Profile
The following are known background and profile answers from this person:
{persona_text}

## Shared Reference Bank
The following are this person's actual answers to related questions. Use the whole bank to calibrate the person's values and response style. Each target question below includes Relevant reference IDs. These IDs point to entries in this Shared Reference Bank and mark the references most semantically related to that target question. For each target question, first use its listed Relevant reference IDs as primary evidence, then use the rest of the bank and the person profile as secondary context:

{fs_block}

## Task
Based on the person profile and shared Reference Bank above, predict this person's answer to each question below.

## Questions
{question_blocks}

# Output Format
Output valid JSON only. Do not wrap in markdown code blocks.
Return exactly one answer object per sample_id. Each answer is a single integer option index, starting from 0.
{"answers": [{"sample_id": "0", "answer": 0}]}

[The {fs_block} placeholder is replaced with:]
### Reference Entries
### Reference {local_idx} (reference_id: "{local_idx}")
Question: {question_text}

Options:
{options_text}

Person's actual answer: {answer}

[The {question_blocks} placeholder contains easy-to-hard target questions in this format:]
### Question {local_idx} (sample_id: "{local_idx}")
Relevant reference IDs: [{ref_ids}]
Question: {question_text}

Options:
{options_text}
\end{PromptBox}

For target question $q_i$, \texttt{Relevant reference IDs} lists the entries in $S_k\cap R_i$ by decreasing target--reference similarity (Section~\ref{subsec:batching}). The complete shared bank remains available as context.

\end{document}